# Densely Semantic Enhancement for Domain Adaptive Region-free Detectors

Bo Zhang, Tao Chen, Bin Wang, *Senior Member, IEEE*, Xiaofeng Wu, Liming Zhang, and Jiayuan Fan

*Abstract*—Unsupervised domain adaptive object detection aims to adapt a well-trained detector from its original source domain with rich labeled data to a new target domain with unlabeled data. Previous works focus on improving the domain adaptability of region-based detectors, *e.g.*, Faster-RCNN, through matching cross-domain instance-level features that are explicitly extracted from a region proposal network (RPN). However, this is unsuitable for region-free detectors such as single shot detector (SSD), which perform a dense prediction from all possible locations in an image and do not have the RPN to encode such instance-level features. As a result, they fail to align important image regions and crucial instance-level features between the domains of region-free detectors. In this work, we propose an adversarial module, namely, densely semantic enhancement module (DSEM), to strengthen the cross-domain matching of instance-level features for region-free detectors. Firstly, to emphasize the important regions of image, the DSEM learns to predict a transferable foreground enhancement mask that can be utilized to suppress the background disturbance in an image. Secondly, considering that region-free detectors recognize objects of different scales using multi-layer feature maps, the DSEM encodes multi-scale representations across different domains. Finally, the DSEM is pluggable into different region-free detectors, ultimately achieving the densely semantic feature matching via adversarial learning. Extensive experiments have been conducted on PASCAL VOC, Clipart, Comic, Watercolor, and FoggyCityscape benchmarks, and their results well demonstrate that the proposed approach not only improves the domain adaptability of region-free detectors but also outperforms existing domain adaptive region-based detectors under various domain shift settings.

## I. INTRODUCTION

OBJECT detection task aims to identify and localize multiple instances of interest in an image, which plays a crucial role in a wide scope of applications, such as intelligent surveillance, instance segmentation, and person reidentification [1]-[6]. Benefitting from the rapid development of deep neural networks (DNNs) in the computer vision community, the ability to learn a robust detector in the large-scale annotated training dataset has been pushed forward a lot [7], [8].

Recently, DNNs based object detection approaches have been extensively studied and can be categorized into region-based (two-stage) detectors [9]-[12] and region-free (one-stage) detectors [13]-[18]. Region-based detectors first generate multiple proposal regions from the input image and then refine these proposal regions through the subsequent classification and position regression heads [10], [11]. However, region of interest (ROI) pooling module [19] in the region-based detector family is designed to extract fixed-length features from different sizes of proposal regions but cannot smoothly propagate the object instance-level gradient to the backbone network [20]. By generating more high-quality proposal regions that cover multiangle objects with rotation variations [21] and better envelop the entire objects [22], the detection performance can be further boosted. Besides, explicitly embedding the rotation invariant regularization and the Fisher discrimination criterion into the objective functions of detection framework is proposed to learn more powerful representations [23]. On the other hand, region-free detectors, including YOLO [13], SSD [14], DSSD [15], RefineDet [16] YOLO9000 [17], DES [18], detect objects directly from the input image in a single stage, saving the ROI pooling and leading to significant speed-up in gradient backpropagation and forward inference. Although these works are inspiring and have achieved promising performance gains on some common benchmarks, they still face major challenges for cross-domain applications due to the difference in data distribution between source and target domains. Several representative cross-domain scenarios are illustrated in Fig. 1, where considerable domain shifts can be observed.

One solution to alleviate the performance degradation problem caused by the data distribution difference between domains is unsupervised domain adaptation (UDA) [24], [25], whose goal is to align the feature distribution by reducing the distribution discrepancy in the feature space using two main technologies, including moment matching [27]-[30] and adversarial learning [31]-[37]. The moment matching approaches refer to aligning the first-order moments and the second-order moments of two different feature distributions using maximum mean discrepancy (MMD) [27]-[29] and deep correlation alignment (CORAL) [30], respectively. Another choice for relieving the cross-domain feature discrepancy is adversarial learning between a backbone network and a domain

Manuscript received November 13, 2020; revised January 26, 2021 and February 25, 2021; accepted March 22, 2021. This work was supported by the National Natural Science Foundation of China under Grant 61731021 and Grant 61971141. This paper was recommended by Associate Editor Yi Yang. (*Corresponding author: Bin Wang*)

B. Zhang, T. Chen, B. Wang, X. Wu, and L. Zhang are with the Key Laboratory for Information Science of Electromagnetic Waves (MoE), Fudan University, Shanghai 200433, China, and also with the Research Center of Smart Networks and Systems, School of Information Science and Technology, Fudan University, Shanghai 200433, China (e-mail: wangbin@fudan.edu.cn).

J. Fan is with the Academy for Engineering and Technology, Fudan University, Shanghai 200433, China.

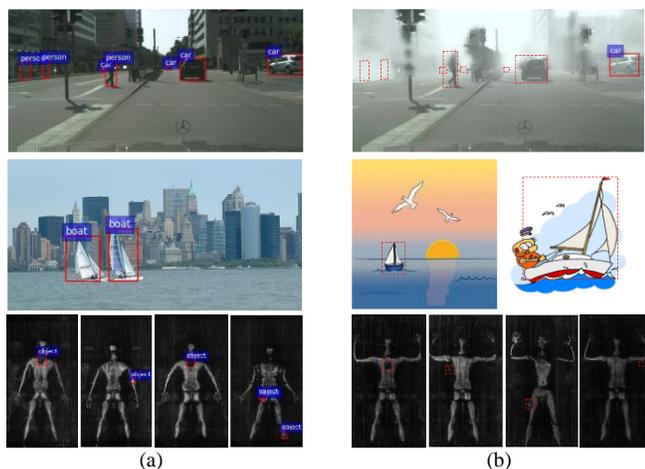

Fig. 1. Some representative cross-domain scenarios. (a) Images from the source domain, where red bounding boxes represent the object annotations. (b) Images from the target domain, where red bounding boxes with dash lines represent the objects that cannot be recognized by the model trained on the source domain data.

discriminative network [31]. Common adversarial learning methods constitute the two networks in a two-player game: the backbone network (generator) that learns discriminative representations from rich labeled source data, and the domain discriminative network that predicts the domain labels given both the labeled source data and unlabeled target data. Besides, the contrastive adaptation network (CAN) [38] is proposed to extend UDA from single-source to multi-source domain adaptation scenarios. Through reinforcement learning (RL) selection strategy, the target validation set is utilized as the reward to train the model [39], [40]. However, it should be noted that these methods are mainly devised on basic image classification task, where only one instance object exists in the whole input image, resulting in the over-emphasis of aligning global semantic features and the neglect of matching semantic features of local instances [41]. As a result, it is intractable and impractical to directly apply these classification based domain adaptation methods to the cross-domain object detection task that is required to consider multiple local instances for feature alignment.

Driven by the UDA in the image classification task, some unsupervised domain adaptive object detection methods [42]-[52] have recently been proposed to deal with domain shifts in the wild, which can be roughly categorized into pixel-level adaptation [42]-[45] and feature-level adaptation [46]-[52]. Pixel-level adaptation methods generate new images following the data distribution of the target domain by an image-to-image translation model [53], and then, they retrain the cross-domain detector using these generated new images. Such an adaptation strategy can appropriately alleviate the distribution discrepancy between domains, on condition that the generated images have good quality and strictly conform to the real distribution of the target domain. However, this is difficult to be guaranteed, as the training of GAN [42] often suffers from mode collapse. Moreover, the training of the image-to-image translation model imposes a heavy computational burden on the rapid deployment of domain adaptive detector in the target domain.

On the other hand, feature-level adaptation detection methods [46]-[52] aim to directly align features from the backbone network [52] or proposal regions [48] such that they can be safely transferred under different real scenarios with domain shifts, which is closely related to the cross-domain detection goal and the cross-domain adaptation process is relatively simpler compared with that of the pixel-level adaptation methods. Inspired by the traditional gradient reversal layer (GRL) module that is designed in order to mimic the adversarial learning process in an end-to-end way [31], an initial attempt of cross-domain detection integrates some adversarial domain classifiers into the Faster RCNN [52] to achieve the feature-level adaptation. Such an attempt is extended by multiple hierarchical adaptation modules [49], strong and weak alignments of features [51], categorical regularization [48] to further boost the domain adaptability of DNNs based detectors. One crucial reason for the breakthrough progress achieved by the above works is that some important local regions in an image are effectively aligned, on condition that region proposal network (RPN) in Faster RCNN can be utilized to explicitly encode instance-level features. The RPN is an important guidance for an outstanding cross-domain region-based detection model and can avoid the redundant adaptation for a large number of background regions. However, for the one-stage region-free detector family having no RPN to encode the instance-level features, such an adaptation strategy of aligning features by means of the RPN encoding process is unreasonable. Further, when these domain adaptation modules, originally designed to meet the requirements of region-based detector architecture, are applied to region-free detectors such as SSD, there are still two major challenges as follows.

Firstly, unlike region-based detectors employing the RPN to produce proposal regions of multiple instances from an input image, region-free detectors first produce spatially-dense feature representations without the encoding process of instance-specific features, and then predict the position offsets and category information according to the spatially-dense representations. However, these dense representations often contain extensive background information. As a result, aligning the dense representations between domains becomes difficult, due to large variations in background appearance and scene layout.

Secondly, most region-free detectors recognize objects with different scales using multi-layer feature maps. Besides, the input image contains complex multi-scale information. For these reasons, the cross-domain adaptation should fully consider the matching of features from different semantic levels and multiple scales.

Accordingly, it is important to combine both the background suppression of spatially-dense representations and the matching of multi-scale representations to perform domain adaptive region-free object detection. In this work, we propose a novel adversarial domain adaptation module, namely, densely semantic enhancement module (DSEM). The DSEM, which can be easily integrated into a variety of object detection networks including SSD, RefineDet, *etc.*, is an end-to-end trainable module for adapting a well-trained detector from its source domain to an unlabeled target domain. Specifically, to sufficiently suppress the background disturbance before the domain adaptation, a foreground mask is designed, which enhances the instance-level information by a semantic segmentation branch. However, due to the annotations from the target domain are unavailable, the semantic segmentation branch trained from the source domain data will result in the prediction deviation for the foreground mask of the target domain. Thus, a domain adaptive foreground enhancement (DAFE) is further developed to produce the transferable foreground mask. Then, the features processed by the above mask are fed into the subsequent domain-related encoder to extract multi-scale features. Finally, with the aid of adversarial learning between the backbone network and the DSEMs, a domain adaptive region-free detection framework is developed, which can sufficiently match the cross-domain features from multi-level semantics.

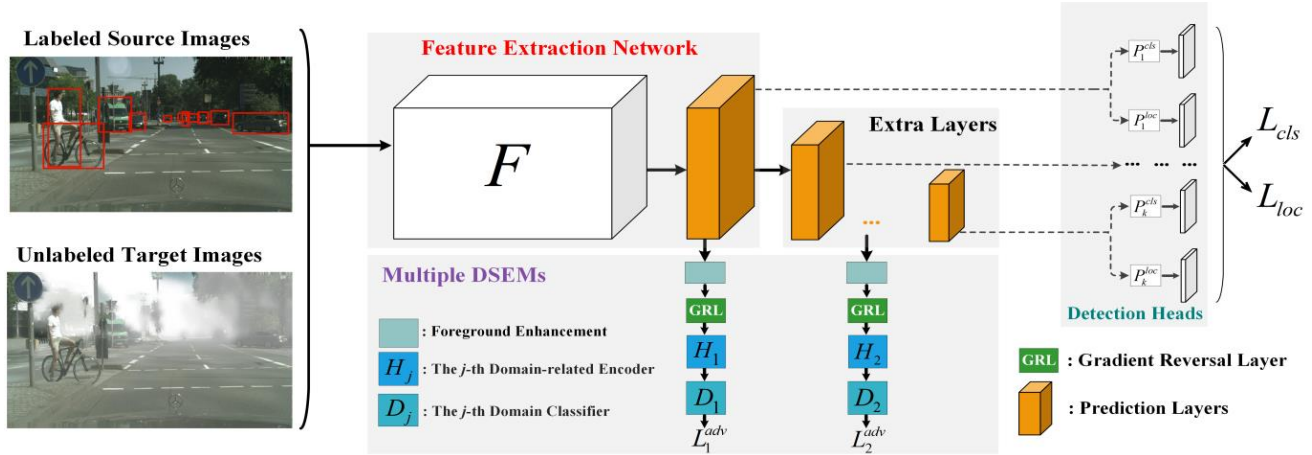

Fig. 2. The overview of the developed domain adaptive region-free detector, which is mainly composed of a feature extraction network, multiple detection heads and multiple DSEMs. Here, SSD [14] is selected as the baseline detector. Our implementation aligns the Conv4_3 and the Conv6_2 layers of SSD. Red bounding boxes in the source images represent the object annotations.

To demonstrate that our proposed module can be generalized to different detectors, we take SSD [14] and RefineDet [16] as examples, plugging the DSEM into these region-free detectors for domain adaptive one-stage detection, and conduct extensive experiments under various domain shift settings, including different image styles [47], weather conditions [54], [55] and human poses [56]. The empirical results and insightful analyses demonstrate that the DSEM not only can sufficiently capture important regions in an image to achieve the alignment process, boosting the adaptability of region-free detectors, but also achieves superior results over other types of object detection models such as region-based detectors. Additionally, we conduct experiments to inspect the effectiveness of DSEM on Faster RCNN [11].

The main contributions of this paper can be briefly summarized as follows:

1) We reveal a crucial aspect of realizing domain adaptive region-free detectors, namely, the cross-domain consistency of multi-scale representations, and propose a domain adaptation method for object detection, which is a pioneer work to adapt the region-free detectors to a new domain.
2) We develop a domain adaptation module, namely DSEM, to suppress the background disturbance and encode domain-related multi-level semantics for feature alignment, which can be flexibly applied to various region-free detectors.
3) We conduct extensive experiments under various cross-domain scenarios, where our method outperforms existing detectors. Besides, the experimental results for real-world applications further demonstrate the generalization ability of our method.

The remainder of this paper is organized as follows: Section II first introduces the problem formulation of cross-domain detection and the detection strategy of region-free detectors. Then, the DSEM is presented in detail. Finally, the optimization objectives and adaptation strategy of the proposed domain adaptive one-stage detection approach are given. In Section III, we evaluate the proposed method on different cross-domain scenarios and various types of detection models. The insightful analyses and discussions, including ablation studies, Faster RCNN with DSEM, sensitivity analyses and model interpretability, and visual examples, are given in Section IV. Concluding remarks are presented in Section V.

## II. THE PROPOSED METHOD

The purpose of this work is to adapt a pre-trained detector from its original source domain to an unlabeled target domain under different cross-domain scenarios. Fig. 2 shows an overall framework of our designed domain adaptive region-free detector, consisting of three parts: a feature extraction network, detection heads in multiple layers, and multiple DSEMs. To better illustrate the domain adaptive detection process coupled with region-free detectors, we first give the problem formulation of cross-domain object detection and the detection strategy of region-free detectors. Then, we present the details of the DSEM. Finally, we give the overall optimization objectives and adaptation strategy of our designed domain adaptive region-free detector.

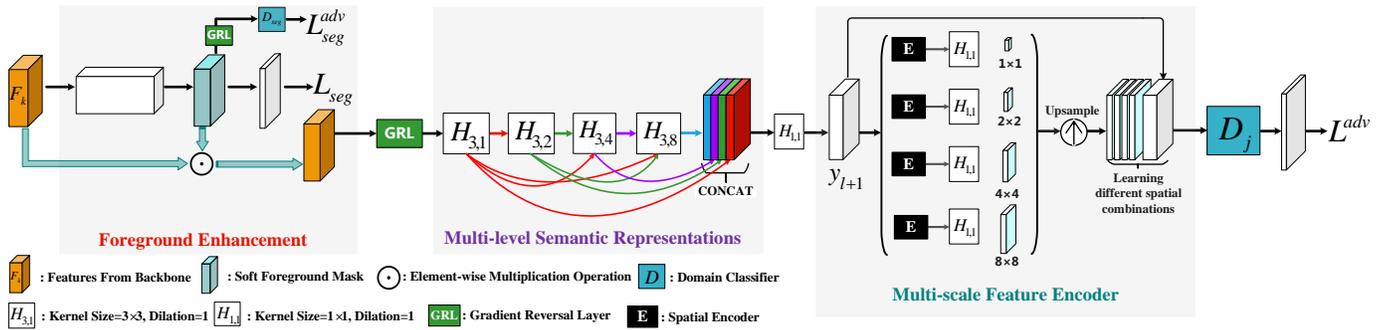

Fig. 3. The illustration of the DSEM. Note that the domain discrepancy in the foreground enhancement can also be reduced *via* an adversarial loss $L_{seg}^{adv}$. Our implementation sets $l$ in Eq. (4) to 3. Gradient Reversal Layer (GRL) in [31] is employed to simply implement adversarial training process.

## A. Preliminaries

*1) Problem Formulation:* Suppose that $F$ is a feature extraction network of an object detection model, and $\{X^s\}$ is the set of all input images from the source domain $s$, and $\{Y^s, B^s\}$ denotes the set of annotations for the corresponding input images where $Y^s$ and $B^s$ represent the category and position labels, respectively. The purpose of a cross-domain detection model is to learn a generalized $F$, such that the $F$ can be adapted to a new target domain $t$, where unlabeled training images $\{X^t\}$ from the target domain are assumed to be available during the adaptation process.

*2) One-stage Region-free Detection Strategy:* Current one-stage region-free detection models are generally composed of a feature extraction network $F$ and some detection heads $P$, where the former adopts common DNNs, *e.g.*, VGG or ResNet, to learn effective representations for detecting objects while the latter is to predict the object category and position information according to the representations learned from the $F$. Most region-free detection models such as SSD [14] and RefineDet [16], employ multi-layer features to predict multiple instances with different scales in an image. Given the *i*-th training image $x_i^s \in X^s$ from the source domain and the *k*-th layer's features extracted by $F_k$, the detection objective is to optimize the $F$ and the $P$ such that the loss function $L_{det}$ is minimized for the source domain data and can be formulated as follows:

$$L_{det}(F,P) = \frac{1}{n_s}\sum_i L_{cls}([P_1^{cls}(F_1(x_i^s)),...,P_k^{cls}(F_k(x_i^s))], Y_i^s) \quad (1)$$
$$+ L_{loc}([P_1^{loc}(F_1(x_i^s)),...,P_k^{loc}(F_k(x_i^s))], B_i^s)$$

where the *k*-th layer's features $F_k(x_i)$ are transformed through the category detection head $P_k^{cls}$ and the position detection head $P_k^{loc}$, and $n_s$ denotes the number of input images from the source domain, and $[...,...]$ represents the concatenation operation, and $L_{cls}$ and $L_{loc}$ are the classification loss and localization loss for multiple instances in an image, respectively.

## B. Densely Semantic Enhancement Module

For easy understanding, we use the typical region-free detector SSD as the baseline detector, and further present the DSEM which includes three parts: domain adaptive foreground enhancement, multi-level semantic representations, and multi-scale feature encoder. The overview of the DSEM is shown in Fig. 3.

*1) Adversarial Learning for Domain Adaptation:* Adversarial learning brings the new driving force to further progress in domain adaptation, where a domain discriminative network is required to accurately predict the domain labels, and a feature extraction network to be aligned is trained to learn transferable features indistinguishable to the domain discriminative network. The cross-domain feature discrepancy is bound to be reduced when the learned feature extraction network successfully deceives the domain discriminative network. The proposed DSEM acts as the domain discriminative network, which will be detailed in the following sections.

*2) Domain Adaptive Foreground Enhancement:* One of the functions achieved by the DSEM is foreground enhancement, which aims to produce representations dominated by foreground instances rather than complex backgrounds for feature alignment. The DSEM can provide a foreground mask to enhance the cross-domain instance-level alignment through a semantic segmentation branch. Note that the semantic segmentation branch is required to suppress the background regions of features during the model adaptation stage, and thus it can be removed during the model inference stage.

Specifically, suppose that $F_k(x_i) \in \mathbb{R}^{C \times U \times V}$ denotes features from a certain layer to be aligned between domains, where $C$, $U$, and $V$ denote the channel number, width, and height of the feature maps, respectively. We expect that before aligning $F_k(x_i)$ between domains, the foreground representations in $F_k(x_i)$ can be enhanced. Specifically, given the foreground annotations $\{C^s, B^s\}$ from the source domain, we first can train a semantic segmentation branch $F_{seg}$ as follows:

$$L_{seg}(F_{seg}) = \log P\left(\tilde{B}_i^s \mid F_{seg}(M_{seg}(F_k(x_i^s)))\right) \quad (2)$$

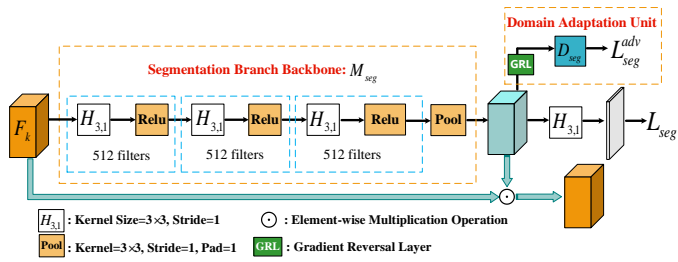

Fig. 4. The illustration of the semantic segmentation branch $F_{seg}$ in DSEM.

where $\tilde{B}^s$ denotes the segmentation level annotations converted from the corresponding bounding-box level annotations, $M_{seg}(F_k(x_i)) \in \mathbb{R}^{C \times U \times V}$ denotes the high-level features from the $F_{seg}$, and $L_{seg}$ adopts a softmax function which is a common objective function in semantic segmentation. The $M_{seg}(F_k(x))$ is defined as a soft foreground mask in the proposed DSEM.

However, it can be observed from Eq. (2) that the $F_{seg}$ is trained by the object annotations from the source domain, which can also be affected by the domain shift. The branch $F_{seg}$ can enhance the foreground information of backbone detection features, provided that the learned foreground mask $M_{seg}(F_k(x))$ in $F_{seg}$ is domain-invariant. For this reason, we design a domain adaptive foreground enhancement by introducing a domain adaptation unit (DA). Specifically, a multi-layer perceptron (MLP) domain classifier $D_{seg}$ with GRL [31] is inserted into $F_{seg}$ to reduce the domain discrepancy in $F_{seg}$. The adversarial objective function $L_{seg}^{adv}$ for the segmentation branch $F_{seg}$ can be written as follows:

$$L_{seg}^{adv}(D_{seg}) = -\frac{1}{n_s UV} \sum_i \sum_u \sum_v d_i \log D_{seg}(M_{seg}(F_k(x_i^s))^{(u,v)}) \\ -\frac{1}{n_t UV} \sum_i \sum_u \sum_v (1-d_i) \log\left(1 - D_{seg}(M_{seg}(F_k(x_i^t))^{(u,v)})\right) \quad (3)$$

where $D_{seg}$ represents a domain classifier that is inserted into the segmentation branch $F_{seg}$ with $d_i = 1$ and $d_i = 0$ for the $i$-th image from the source domain and the target domain, respectively, and $M_{seg}(F_k(x_i))^{(u,v)}$ is the activation value located at $(u,v)$ of the feature map. $n_t$ denotes the number of input images from the target domain.

Further, the domain adaptive foreground enhancement can be formulated by an element-wise multiplication operation $M_s \odot F_k(x_i^s)$ for the source features or $M_t \odot F_k(x_i^t)$ for the target features, where $M_s = M_{seg}(F_k(x_i^s))$ and $M_t = M_{seg}(F_k(x_i^t))$ represent the domain adaptive foreground mask of features. The visual network structure of $F_{seg}$ is illustrated in Fig. 4.

With the aid of the domain adaptive foreground enhancement, we can appropriately suppress the redundant background information of features from the backbone, and the processed features will be fed into the subsequent two parts for encoding multi-scale features as follows.

*3) Multi-level Semantic Representations:* In the existing adversarial domain adaptive object detection approaches [48]-[52], the domain discriminative network is typically composed of a domain-related encoder and a domain classifier. The former produces the domain-related representations while the latter predicts the domain labels based on these representations. The domain-related encoder is indispensable since it indicates which features are important to distinguish the source domain from the target domain. Subsequently, these selected features are fully aligned *via* adversarial learning. However, for the region-free detectors using multi-layer features for prediction such as SSD [14] and RefineDet [16], the cross-domain discrepancy could reside in multiple prediction layers, increasing the risk of semantic inconsistency in multiple scales or layers.

Accordingly, for the domain-related encoder, fusing the instance-level (foreground) features extracted from the multiple layers of the feature extraction network is crucial. In order to achieve this aim, we employ a stack of dilated convolutional layers connecting in a dense way, for emphasizing the learning of domain-related dense semantic representations.

In particular, let $H_{k,d}(\cdot)$ denote a convolution transformation with kernel size $k \times k$ and dilation rate $d$. As our domain discriminative network adopts a densely-connected way, we employ $y_0 = H_{3,1}(M \odot F_k(x_i))$ to reduce the number of channels of feature maps from the feature extraction network. After that, the input feature maps for a certain layer $l$ are concatenated with all the outputs from lower layers, and fed into the $l$-th dilation convolution layer as follows:

$$y_l = H_{3,2^l}([y_{l-1}, y_{l-2}, ..., y_0]) \quad (4)$$

where $y_l$ represents the output of the $l$-th layer. To consider densely semantic information across all layers, we concatenate the output of every layer of domain-related representations, and finally reduce the number of channels after the concatenation operation by $H_{1,1}(\cdot)$, which can be written as:

$$y_{l+1} = H_{1,1}([y_l, y_{l-1}, ..., y_0]) \quad (5)$$

where $l$ is set to 3 since we have observed that from the experiments, the modeling power of the atrous convolution gradually deteriorates when $l$ goes beyond 4.

*4) Multi-scale Feature Encoder:* On the other hand, DSEM needs to consider spatial combination relationships across multi-scale features. We can easily encode the multi-scale information for features $y_{l+1}$, provided that the redundant background information in the $y_{l+1}$ has been appropriately weakened by the above domain adaptive foreground enhancement and the densely semantic information from different levels has been encoded into the $y_{l+1}$.

Specifically, since the pooling operation with kernels of different sizes can capture different sub-regions in feature maps, we feed $y_{l+1}$ into multiple pooling branches to extract features with different spatial sizes. Then, their spatial combination relationships can be explicitly learned in an end-to-end way by a convolution transformation with kernel size $1\times 1$. As shown in Fig. 3, our designed pooling branches adopt four-level pyramid pooling pathways with bin sizes of $1\times 1$, $2\times 2$, $4\times 4$ and $8\times 8$ at each spatial level, and then, they apply convolutional layers with $1\times 1$ kernel-size to learn the combination relationships between different sub-regions. It is worth noting that the ultimately produced features will be aligned in a spatially-dense way, since we find that the spatially-dense alignment is conducive to preserving more spatial details. Moreover, some recent works [49]-[51], *e.g.*, weak global alignment, convert the feature maps into a single vector *via* a global average pooling (GAP). This type of pooling operation that ignores spatial relationships and enhances global representations is effective for cross-domain image classification, but not enough for cross-domain detection which requires more spatial distribution information of instances. In contrast to the above methods, our design can be considered as a more general form of the above-mentioned GAP operation.

With the above definition, we can encode densely semantic features and spatial relationships. Let $H_j(M \odot F_k(x)) \in R^{C\times U\times V}$ denote the final output of the $j$-th domain-related encoder in DSEM. The $j$-th adversarial objective $L_j^{adv}$ of the DSEM can then be written as follows:

$$L_j^{adv}(H_j, D_j) = -\frac{1}{n_s UV}\sum_i\sum_u\sum_v d_i \log D_j(H_j(M \odot F_k(x_i^s)))^{(u,v)}$$
$$-\frac{1}{n_t UV}\sum_i\sum_u\sum_v (1-d_i)\log(1-D_j(H_j(M \odot F_k(x_i^t)))^{(u,v)}) \quad (6)$$

where $D_j$ represents the $j$-th domain classifier, and $H(M \odot F(\cdot))^{(u,v)}$ is the activation value located at $(u,v)$ of the feature map.

### C. Overall Optimization Objectives and Adaptation Strategy

*1) Optimization Objectives:* In order to match hierarchical features between domains, the DSEM is simultaneously inserted into multiple prediction layers when performing the cross-domain adaptation phase. The loss function of the developed domain adaptive region-free detector can be written as follows:

$$\min_{F,P} L_{det}(F, P) \quad (7)$$

$$\max_{D_{seg}} \min_{F_{seg}} L_{seg}(F_{seg}) - L_{seg}^{adv}(D_{seg}) \quad (8)$$

$$\max_{H,D} \min_{F,P} L_{det}(F, P) - \lambda \sum_{j=1}^{n_j} L_j^{adv}(H_j, D_j) \quad (9)$$

where $n_j$ denotes the number of the DSEM inserted into region-free detectors, and the $P$ represents a combination of the $P^{cls}$ and the $P^{loc}$, and $\lambda$ denotes a hyper-parameter that can trade off the importance between the model discriminability for source data and the model adaptability for target data.

*2) Adaptation Strategy:* We consider the following two-stage adaptation strategy to perform the cross-domain alignment of features. Firstly, to ensure that the detection model can learn sufficient knowledge for the subsequent model transfer, a source model is pre-trained using the annotated images from the source domain by Eq. (7). Secondly, the adversarial adaptation phase is then performed by fine-tuning the above pre-trained detection model by Eqs. (8) and (9) on the mixture set which is composed of annotated source images and unannotated target images. Note that the semantic segmentation branch $F_{seg}$ is integrated into the baseline detection model along with the DSEM only in the adversarial adaptation phase. In this phase, the sign of gradients is flipped by GRL proposed in [31]. Compared with the optimization strategy employed in the existing works [48]-[52] that only consider the cross-domain matching of instance-level features from Faster RCNN, we simultaneously consider the cross-domain foreground enhancement that can be expressed by Eq. (8) and the cross-domain matching of densely semantic representations and multi-scale features that can be expressed by Eq. (9).

After the cross-domain adaptation process ends, the DSEM can be removed during the model inference phase. Hence, the developed domain adaptive region-free detector will not bring additional computational costs to the original one for testing use.

## III. EXPERIMENTS

In this section, we firstly introduce the details of datasets used to evaluate our proposed approach. Then, we give the experimental setup for the pre-training and adaptation phases. Further, we evaluate the proposed approach under two typical cross-domain scenarios, including different image styles, weather conditions. Finally, we apply the proposed approach to real-world application scenarios to further demonstrate its generalization capability.

### A. Dataset

Six public datasets are utilized to evaluate our method, including PASCAL VOC [57], Clipart [47], Comic [47], Watercolor [47], Cityscapes [54], and FoggyCityscapes [55].

*1) PASCAL VOC:* This dataset represents real-world images containing 20 categories. We follow the official split that is widely used in [14] and [16], which adopts both the *trainval* list on PASCAL VOC 2007 (containing 5011 images) and the *trainval* list on PASCAL VOC 2012 (containing 11540 images) as the train set, and employs *test* list on PASCAL VOC (containing 4952 images) as the test set.

*2) Clipart, Comic, and Watercolor:* These datasets represent graphical images, cartoon images, and painting, respectively, where these artistic or hand-painted images have large domain gaps with real-world images. For Comic and Watercolor, the official *train* list (including 1000 images) is used as the train set, and the official *test* list (including 1000 images) is utilized as the test set.

TABLE I
RESULTS ON ADAPTATION FROM PASCAL VOC TO CLIPART.
G and L denote weakly global alignment and strongly local alignment in G-L-Faster [51], respectively. LW-SSD performs the layer-wise alignment from conv4_3 to conv9_2 using the G and L. DN-8 and DN-32 denote that the DSEM is used to align the features that are down-sampled 8 times and 32 times relative to the input images, respectively. P-A represents pixel-level adaptation, which generates extra annotated target images by CycleGAN [53]. RDet denotes RefineDet [16]. "*" indicates that we employ the COCO pre-trained model and fine-tune on VOC 07+12. "♠" and "♣" represent models using VGG-16 and ResNet-101 backbones, respectively. Note that all our proposed models adopts the VGG-16. The best results based on the ImageNet pre-trained model are shown in bold font.

|  | Method | G | L | DN-8 | DN-32 | P-A | aero | bcy. | bird | boat | bott. | bus | car | cat | chair | cow | table | dog | hrs | bike | prsn | plnt | sleep | sofa | train | tv | mAP |
|---|---|---|---|---|---|---|---|---|---|---|---|---|---|---|---|---|---|---|---|---|---|---|---|---|---|---|---|
| Region-free | SSD [14] | | | | | | 20.9 | 56.2 | 20.3 | 16.4 | 9.5 | 38.1 | 33.9 | 10.9 | 37.6 | 22.9 | 22.6 | 10.6 | 22.6 | 48.9 | 43.3 | 35.2 | 7.3 | 30.2 | 36.2 | 27.5 | 27.6 |
| | L-SSD | | √ | | | | 20.8 | 58.2 | 19.2 | 17.4 | 11.9 | 47.1 | 38.6 | 11.0 | 35.4 | 22.9 | 22.9 | 16.2 | 23.5 | 50.3 | 45.6 | 36.6 | 9.5 | 33.3 | 39.0 | 31.1 | 29.5 |
| | G-SSD | √ | | | | | 19.6 | 55.3 | 21.5 | 19.8 | 8.1 | 45.9 | 32.1 | 6.9 | 37.1 | 22.7 | 26.1 | 10.6 | 24.8 | 59.5 | 45.6 | 34.5 | 11.8 | 34.5 | 41.2 | 32.1 | 29.5 |
| | G-L-SSD | √ | √ | | | | 20.7 | 62.2 | 21.6 | 22.7 | 20.4 | 44.8 | 34.8 | 9.4 | 38.8 | 24.7 | 26.5 | 10.6 | 22.7 | 64.1 | 48.6 | 36.1 | 10.2 | 32.4 | 43.9 | 34.6 | 31.5 |
| | LW-SSD | √ | √ | | | | 19.8 | 48.4 | 27.4 | 26.2 | 25.1 | 62.2 | 40.0 | 7.3 | 38.9 | 38.4 | 25.1 | 7.7 | 14.1 | 65.6 | 53.4 | 41.5 | 14.1 | 34.4 | 49.6 | 48.5 | 34.3 |
| | SSD+DSEMs (ours) | | | √ | | | 20.5 | 58.9 | 29.1 | 27.1 | 27.3 | 54.5 | 39.1 | 11.3 | 40.9 | 42.5 | 30.2 | 12.9 | **29.3** | 75.2 | 56.9 | 45.3 | 16.1 | **39.6** | 57.3 | **48.9** | 38.1 |
| | | | | √ | √ | | 23.8 | **63.8** | 27.3 | 27.9 | **31.2** | 60.5 | 41.2 | **16.7** | 45.2 | 47.7 | 38.6 | **16.3** | 27.4 | **77.6** | 58.2 | **49.2** | 17.1 | 31.5 | 50.9 | 48.2 | 40.1 |
| | | | | √ | √ | √ | **25.9** | 62.3 | **30.1** | **34.2** | 27.0 | **77.4** | **47.2** | 12.2 | **45.9** | **48.8** | **40.1** | 11.8 | 28.0 | 75.5 | **62.8** | 43.4 | **23.7** | 37.7 | **61.9** | 48.4 | **42.2** |
| | RDet [16] | | | | | | 20.0 | 41.5 | 21.7 | 17.5 | 25.8 | 46.2 | 24.0 | 10.9 | 34.7 | 12.5 | 24.5 | 16.2 | 17.9 | 48.8 | 32.3 | 39.8 | 3.0 | 20.3 | 35.0 | 26.6 | 26.0 |
| | RDet+DSEMs (ours) | | | √ | | | 20.3 | 55.0 | 25.1 | 17.5 | **50.7** | 52.5 | 27.3 | 17.5 | 36.4 | 20.5 | 19.4 | 17.9 | 23.0 | 65.3 | 43.6 | 48.1 | 12.6 | 23.6 | 44.4 | 41.6 | 33.1 |
| | | | | √ | √ | | 24.1 | 58.3 | 29.1 | **26.2** | 46.4 | 61.0 | 39.5 | 17.5 | 44.3 | 44.9 | 25.9 | 16.8 | 28.1 | 65.1 | 60.9 | 49.4 | **18.9** | 30.8 | 56.3 | 50.6 | 39.8 |
| | | | | √ | √ | √ | **28.9** | **71.7** | **31.8** | 19.3 | 44.7 | **69.3** | **44.6** | **24.1** | 40.4 | 39.8 | 22.5 | **22.2** | **30.5** | **92.4** | **63.8** | **51.0** | 16.1 | **32.1** | **67.0** | **56.4** | **43.5** |
| | WST+BSR [46] | | | | | | 28.0 | 64.5 | 23.9 | 19.0 | 21.9 | 64.3 | 43.5 | 16.4 | 42.2 | 25.9 | **30.5** | 7.9 | 25.5 | 67.6 | 54.5 | 36.4 | 10.3 | 31.2 | 57.4 | 43.5 | 35.7 |
| | SSD* [14] | | | | | | 23.1 | 59.8 | 23.1 | 15.0 | 18.0 | 58.5 | 40.8 | 15.1 | 41.5 | 40.1 | 33.4 | 20.1 | 29.8 | 58.9 | 49.7 | 25.3 | 18.4 | 30.5 | 41.8 | 43.8 | 34.3 |
| | SSD*+DSEMs | | | √ | √ | | 27.9 | 62.1 | 29.7 | 28.9 | 38.6 | 81.5 | 50.7 | 14.9 | 49.5 | 56.1 | 40.2 | 15.6 | 38.7 | 73.4 | 60.5 | 39.5 | 21.5 | 41.3 | 63.1 | 51.7 | 44.3 |
| | RDet* [16] | | | | | | 26.0 | 55.9 | 28.0 | 25.4 | 34.4 | 52.3 | 45.1 | 16.4 | 52.8 | 25.9 | 26.8 | 19.1 | 40.7 | 50.3 | 46.1 | 41.3 | 16.1 | 29.6 | 47.3 | 32.6 | 35.6 |
| | RDet*+DSEMs | | | √ | √ | | 30.0 | 60.3 | 39.1 | 30.6 | 55.4 | 69.2 | 55.6 | 27.5 | 51.3 | 52.1 | 37.7 | 26.7 | 43.3 | 77.0 | 72.0 | 59.0 | 26.5 | 43.1 | 64.9 | 56.1 | 48.9 |
| Region-based | Faster [11]♠ | | | | | | 15.7 | 31.9 | 22.4 | 8.2 | 38.8 | 59.4 | 17.8 | 6.6 | 37.0 | 5.7 | 12.7 | 7.2 | 17.4 | 49.0 | 36.0 | 32.1 | 11.2 | 2.9 | 29.8 | 28.4 | 23.5 |
| | Faster [11]♣ | | | | | | **35.6** | 52.5 | 24.3 | 23.0 | 20.0 | 43.9 | 32.8 | 10.7 | 30.6 | 11.7 | 13.8 | 6.0 | 36.8 | 45.9 | 48.7 | 41.9 | 16.5 | 7.3 | 22.9 | 32.0 | 27.8 |
| | DA-Faster [52]♣ | √ | | | | | 15.8 | 33.9 | 22.5 | 14.8 | 24.9 | 48.7 | 27.9 | 12.5 | 32.7 | 35.5 | 21.3 | 17.9 | 17.4 | 55.0 | 48.5 | 34.8 | 11.4 | 21.3 | 47.1 | 37.7 | 29.1 |
| | G-L-Faster [51]♠ | √ | √ | | | | 16.0 | 53.2 | 27.5 | 21.6 | 32.0 | 48.4 | 32.4 | 12.2 | 32.5 | 27.3 | 12.3 | 13.1 | 24.3 | 62.4 | 55.5 | 41.2 | 21.0 | 13.2 | 37.8 | 46.1 | 31.5 |
| | G-L-Faster [51]♣ | √ | √ | | | | 26.2 | 48.5 | **32.6** | 33.7 | 38.5 | 54.3 | 37.1 | **18.6** | 34.8 | **58.3** | 17.0 | 12.5 | 33.8 | 65.5 | 61.6 | **52.0** | 9.3 | 24.9 | **54.1** | 49.1 | 38.1 |
| | ICR-CCR [48]♣ | | | | | | 28.7 | 55.3 | 31.8 | 26.0 | 40.1 | **63.6** | 36.6 | 9.4 | 38.7 | 49.3 | 17.6 | 14.1 | 33.3 | 74.3 | 61.3 | 46.3 | **22.3** | 24.3 | 49.1 | 44.3 | 38.3 |
| | DD+MRL [45]♣ | √ | √ | | | √ | 25.8 | **63.2** | 24.5 | **42.4** | **47.9** | 43.1 | **37.5** | 9.1 | **47.0** | 46.7 | **26.8** | **24.9** | **48.1** | **78.7** | **63.0** | 45.0 | 21.3 | **36.1** | 52.3 | **53.4** | **41.8** |

TABLE II
RESULTS ON VOC 2007 TEST SET ON ADAPTATION FROM PASCAL VOC TO CLIPART. THE DEFINITION OF * FOLLOWS TABLE I

| Method | DN-8 | DN-32 | mAP |
|---|---|---|---|
| SSD | | | **77.5** |
| SSD+DSEMs (ours) | √ | | 76.4 |
| | √ | √ | 76.3 |
| RefineDet | | | **81.7** |
| RDet+DSEMs (ours) | √ | √ | 81.0 |
| SSD* | | | **79.5** |
| SSD*+DSEMs (ours) | √ | √ | 79.1 |
| RDet* | | | **85.1** |
| RDet*+DSEMs (ours) | √ | √ | 84.0 |

*3) Cityscapes:* This dataset aims to collect high-quality images of outdoor street scenes under normal weather conditions, which is divided into 2975 images and 500 images for the train set and validation set, respectively.

*4) FoggyCityscapes:* FoggyCityscapes simulates the foggy weather condition with different visibility ranges. The train/ val splits in this dataset are consistent with the Cityscapes.

B. Experimental Setup

To make a comprehensive comparison with current works, different region-free detectors including SSD [14] and RefineDet [16] are selected as the baseline detector.

*1) Pre-training Phase:* Following the training setting in [14] and [16], the input images are resized to 300×300 pixels for SSD [14] and 512×512 pixels for RefineDet [16], respectively. We train the detection network initialized from the ImageNet pre-trained model on the train set from the source domain. Specifically, an initial learning rate of 0.001 is employed for the first $80k$ iterations, and the learning rate decays to 0.0001 and 0.00001 at $100k$ and $120k$ iterations, respectively.

*2) Adversarial Adaptation Phase:* For this phase, we insert multiple DSEMs into the detection network, and fine-tune the model pre-trained by the above phase on the train set of labeled source domain and unlabeled target domain. We adopt a batch size of 16, a momentum of 0.9 and a weight decay of 0.0005. Besides, the initial learning rate is set to 0.0001 in this phase, and the learning rate of the DSEM is set at ten times of the detection network. The overall training of the adaptation phase is finished when 5000 iterations are reached.

*3) Pixel-level Adaptation Phase:* To perform the pixel-level adaptation, we train CycleGAN [53] for 20 epochs with 0.00001 learning rate in the first 10 epochs and a linear decaying rate to zero in the last 10 epochs. Next, we fine-tune the source-domain detector on both the generated target-domain-like images and the original source images.

For fair comparison, we use the results of all region-based and region-free detectors reported in their original paper. For all experiments, we evaluate our method using mean Average Precisions (mAP) under the threshold of 0.5.

TABLE III
RESULTS ON ADAPTATION FROM PASCAL VOC TO COMIC.
THE EVALUATION OF TARGET DOMAIN AND SOURCE DOMAIN IS ON THE TEST SET OF COMIC AND TEST SET OF PASCAL VOC 2007, RESPECTIVELY. THE DEFINITION OF DN-8, DN-32, AND P-A FOLLOWS TABLE I.

| Method | DN-8 | DN-32 | P-A | Target Domain | | | | | | | Source Domain |
|---|---|---|---|---|---|---|---|---|---|---|---|
| | | | | bicycle | bird | car | cat | dog | prsn | mAP | mAP |
| SSD [14] | | | | 21.7 | 12.8 | **34.4** | 11.0 | 14.6 | 44.4 | 23.1 | 81.4 |
| SSD+DSEMs (ours) | ✓ | | | 39.7 | 15.2 | 22.6 | 14.9 | 25.9 | 50.3 | 28.1 | 81.1 |
| | ✓ | ✓ | | 49.6 | 18.2 | 26.6 | **28.8** | 30.8 | 46.3 | 33.4 | 80.1 |
| | ✓ | ✓ | ✓ | **57.8** | **22.2** | 32.2 | 28.5 | **32.9** | **56.8** | **38.4** | 79.5 |
| ADDA [32] | | | | 39.5 | 9.8 | 17.2 | 12.7 | 20.4 | 43.3 | 23.8 | \ |
| DD+MRL [45] | | | | \ | \ | \ | \ | \ | \ | 34.5 | \ |
| WST+BSR [46] | | | | 50.6 | 13.6 | 31.0 | 7.5 | 16.4 | 41.4 | 26.8 | \ |
| DT [47] | | | | 43.6 | 13.6 | 30.2 | **16.0** | 26.9 | 48.3 | 29.8 | \ |

TABLE IV
RESULTS ON ADAPTATION FROM PASCAL VOC TO WATERCOLOR. ORACLE REFERS TO TRAINING THE DETECTOR ON THE LABELED TARGET IMAGES.

| | Method | bicycle | bird | car | cat | dog | prsn | mAP |
|---|---|---|---|---|---|---|---|---|
| Region-free | SSD [14] | 65.8 | 44.0 | 46.0 | 26.7 | 27.7 | 60.5 | 45.1 |
| | DT [47] | **82.8** | 47.0 | 40.2 | 34.6 | 35.3 | 62.5 | 50.4 |
| | WST+BSR [46] | 75.6 | 45.8 | **49.3** | 34.1 | 30.3 | 64.1 | 49.9 |
| | SSD+DSEMs w/o Pixel-level align. | 80.1 | **52.4** | 44.4 | **35.2** | **40.9** | **72.3** | **54.2** |
| | Oracle | 76.0 | 60.0 | 52.7 | 41.0 | 43.8 | 77.3 | 58.4 |
| Region-based | Faster [11] | 68.8 | 46.8 | 37.2 | 32.7 | 21.3 | 60.7 | 44.6 |
| | DA-Faster [52] | 75.2 | 40.6 | **48.0** | 31.5 | 20.6 | 60.0 | 46.0 |
| | G-L-Faster [51] | 82.3 | 55.9 | 46.5 | **32.7** | 35.5 | 66.7 | 53.3 |
| | DD+MRL [45] | \ | \ | \ | \ | \ | \ | 52.0 |
| | Oracle | 83.6 | 59.4 | 50.7 | 43.7 | 39.5 | 74.5 | 58.6 |

## C. Experimental Results

*1) Adaptation across Different Image Styles:* For this cross-domain scenario, we adopt the PASCAL VOC as the source domain and use the Clipart, Comic, and Watercolor as the target domain, respectively. We not only report the experimental results of our methods on the test set of target domain, but also give the results on the test set of source domain. Note that for adaptation from PASCAL VOC to Clipart, 1000 images from the Clipart are used for both the adversarial adaptation phase (only unlabeled images are given) and the testing phase, which accords with the cross-domain setting in [48], [51] and [52].

As shown in Table I, we perform cross-domain alignment for SSD. By comparing L-SSD, G-SSD, G-L-SSD, LW-SSD and SSD, it can be seen that applying some Faster RCNN based domain adaptation modules [49]-[51] to the SSD detector only achieves limited performance improvement. However, by employing the DSEM to align the conv4_3 layer (DN-8) of SSD, the SSD+DSEM compares favorably against G-L-SSD and LW-SSD, since the cross-domain feature alignment of instances is fully considered. Further, aligning both the conv4_3 and conv6_2 (DN-32) layers increases the mAP value from 38.1% to 40.1%. When we further perform pixel-level adaptation using target-domain-like images generated by a CycleGAN model [53], the mAP value (42.2%) is increased by approximately 2%.

Although the domain adaptive detectors can achieve higher mAPs on a new target domain, their detection accuracies on the original source domain usually degrade after the cross-domain adaptation process is finished [48], [50], [51]. Here, we inspect the impact of the proposed DSEM on the detection accuracy of the source domain. From Table II, it can be seen that the proposed methods (such as SSD+DSEMs, *etc.*) can also maintain the high detection accuracies on the source domain.

To further validate that the DSEM can be generalized to different region-free detectors, we select RefineDet as the new baseline detector and repeat the above experiments. It can be concluded from Tables I and II that the DSEM yields consistent performance improvement on another region-free detector.

Next, we compare the developed domain adaptive region-free detectors with state-of-the-art domain adaptive region-based methods [45], [48], [51], [52]. DD+MRL [45] generates multiple intermediate domain images by CycleGAN based method, and then performs unsupervised multi-domain alignment, which uses both the pixel-level and feature-level adaptation, while DA-Faster [52], G-L-Faster [51], and ICR-CCR [48] only use feature-level adaptation. For fair comparison, we compare the RDet+DSEMs using both pixel-level and feature-level adaptation (43.5%) with DD+MRL (41.8%), and compare the SSD+DSEMs using only feature-level adaptation (40.1%) with G-L-Faster (38.1%) [51] and ICR-CCR (38.3%) [48]. The results demonstrate that the developed domain adaptive region-free detectors consistently outperform existing region-based detectors in terms of mAP value.

Large-scale data is indispensable for improving DNNs based detectors. In this work, we studied the impact of large-scale data on unsupervised domain adaptive detectors. In Table I, SSD* and RDet* refer to results that COCO model replaces ImageNet-pre-trained model to initialize the network weights. It can be seen that employing COCO model has significantly increased the mAP by 6.7~9.6%. In other words, the pre-trained model on large-scale data can significantly relieve the domain discrepancy, increasing the mAP value on the target domain. Further, by utilizing DSEMs, the best mAP value of 48.9% is achieved for RDet.

Furthermore, in Tables III and IV, the work [47] adopts a two-stage progressive fine-tuning, including unsupervised pixel-level adaptation (DT) and weakly-supervised adaptation (PL). For [47], in order to compare the unsupervised adaptation part of this method with our method, we only report their results of DT part. It can be seen from Table III and Table IV that our method greatly exceeds the ADDA [32], DT, DA-Faster and G-L-Faster on the test set of both Comic and Watercolor. Furthermore, the mAP value on the test set of Comic is raised to 38.4% when employing the pixel-level alignment, which also greatly outperforms DD+MRL.

*2) Adaptation across Different Weather Conditions:* A safe driverless system requires that the detector can obtain consistently good results under different weather conditions. Hence, we evaluate our method under the domain shift from normal weather to foggy weather. For model adaptation from Cityscapes [54] to FoggyCityscapes [55], the official train set from Cityscapes is used as the source domain. The official train set from FoggyCityscpaes is used as the target domain, and the results are reported on the *validation* set of FoggyCityscapes. Following [48]-[50], we convert the pixel-level annotations into bounding boxes by calculating the tightest rectangles of its instance masks. For this cross-domain scenario, the adversarial adaptation phase lasts $10k$ iterations, with a learning rate of 0.0001 for the first $5k$ iterations and decreased by a factor of 10 for the next $5k$ iterations.

TABLE V
RESULTS ON THE VALIDATION SET OF FOGGYCITYSCAPES. THE DEFINITION OF G, L, LW, DN-8 AND DN-16 FOLLOWS TABLE I. ORACLE REFERS TO TRAINING THE DETECTOR ON THE LABELED TARGET IMAGES.

| | Method | G | L | DN-8 | DN-16 | bus | bicycle | car | bike | prsn | rider | train | truck | mAP |
|---|---|---|---|---|---|---|---|---|---|---|---|---|---|---|
| Region-free | RDet [16] | | | | | 24.3 | 28.9 | 38.0 | 21.8 | 26.1 | 28.5 | 8.3 | 6.6 | 22.8 |
| | L-RDet | | √ | | | 32.0 | 33.3 | 44.6 | 26.8 | 30.8 | 34.3 | 20.0 | 12.7 | 29.3 |
| | G-RDet | √ | | | | 31.9 | 32.1 | 44.3 | 27.0 | 30.3 | 33.9 | 20.2 | 16.2 | 29.5 |
| | G-L-RDet | √ | √ | | | 33.5 | 33.6 | 46.7 | 30.1 | 32.4 | 34.7 | 25.3 | 15.5 | 31.5 |
| | LW-RDet | √ | √ | | | 33.8 | 32.1 | 45.2 | 26.7 | 31.4 | 35.3 | 25.8 | 14.3 | 30.6 |
| | RDet+DSEMs (ours) | | | √ | | 40.7 | 35.3 | 55.1 | 27.4 | 34.8 | **38.5** | 26.5 | 19.0 | 34.7 |
| | | | | √ | √ | **40.8** | **35.3** | **56.2** | **31.2** | **35.9** | 38.1 | **34.9** | **21.2** | **36.7** |
| | Oracle | | | | | 41.9 | 38.7 | 63.3 | 33.3 | 39.9 | 42.8 | 31.8 | 27.3 | 39.8 |
| Region-based | Faster [11] | | | | | 22.3 | 26.5 | 34.3 | 15.3 | 24.1 | 33.1 | 3.0 | 4.1 | 20.3 |
| | DA-Faster [52] | √ | | | | 25.0 | 31.0 | 40.5 | 22.1 | 35.3 | 20.2 | 20.0 | 27.1 | 27.6 |
| | G-L-Faster [51] | √ | √ | | | 36.2 | 35.3 | 43.5 | 30.0 | 29.9 | 42.3 | 32.6 | 24.5 | 34.3 |
| | MAF [49] | | | | | 39.9 | 33.9 | 43.9 | 29.2 | 28.2 | 39.5 | 33.3 | 23.8 | 34.0 |
| | ICR-CCR [48] | | | | | **45.1** | 34.6 | **49.2** | **30.3** | 32.9 | **43.8** | **36.4** | **27.2** | **37.4** |
| | DD+MRL [45] | | | | | 38.4 | 32.2 | 44.3 | 28.4 | 30.8 | 40.5 | 34.5 | 27.2 | 34.6 |
| | Oracle | | | | | 51.9 | 37.8 | 53.0 | 36.8 | 36.2 | 47.7 | 41.0 | 34.7 | 42.4 |

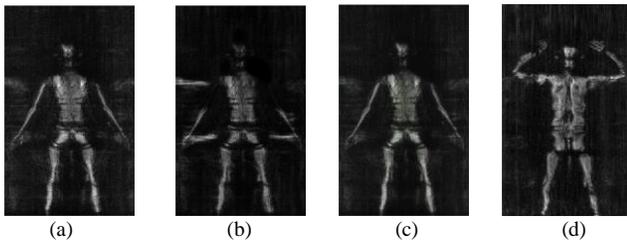

(a) (b) (c) (d)

Fig. 5. Visual results of employing a CycleGAN based model to perform the pixel-level adaptation from the hands-down to hands-up postures. (a) The input image from the source domain. (b) The generated target-domain-like image using CycleGAN [53]. (c) The reconstructed image using CycleGAN [53]. (d) The real target domain image.

As shown in Table V, the performance gain is relatively limited by inserting some typical domain adaptation modules into Rdet, such as local and global alignment modules in [49], [51], [52]. However, the RDet with DSEMs significantly boosts the domain adaptability of the region-free detector. Besides, it should be noted that the experimental results of the RDet with DSEMs (36.7%) are slightly lower than that of ICR-CCR (37.4%), since the ICR-CCR selects a high-performance baseline model (34.8% on FoggyCityscapes) [51]. In contrast, our DSEMs boost the mAP value of RDet from 22.8% to 36.7%. The performance improvement is considerable for the region-free detector family.

### D. Real-world Applications

*1) Cross-posture Hidden Object Detection:* We further apply our approach to a real-world application. Consider this important security inspection scenario: given scanning images for passengers in checkpoints using the millimeter-wave (MMW) imaging technology, the purpose of cross-posture hidden object detection is to locate dangerous items carried by passengers. For the common MMW dataset [56], the standing posture of a human body is divided into two kinds: HD denoting a hands-down posture and HU denoting a hands-up posture, as shown in Figs. 1 and 5. Whether the hidden object detection approach can accurately locate dangerous items under different human body poses is important in real application, since the unconstrained pose for the subjects can make the security process more humanized and convenient.

TABLE VI
FEATURE-LEVEL ADAPTATION ACROSS DIFFERENT POSES ON MMW DATASET. HD AND HU ARE USED AS THE SOURCE DOMAIN AND TARGET DOMAIN, RESPECTIVELY. AVERAGE PRECISION (AP) IS FOR EVALUATION.

| | Method | G | L | DN-8 | DN-32 | HD→HU on HU (IOU=0.5) |
|---|---|---|---|---|---|---|
| Region-free Methods | SSD [14] | | | | | 15.5 |
| | L-SSD | | √ | | | 21.7 |
| | G-SSD | √ | | | | 33.3 |
| | G-L-SSD | √ | √ | | | 35.2 |
| | SSD+DSEMs (ours) | | | √ | | 43.8 |
| | | | | √ | √ | **45.2** |
| | Oracle | | | | | 61.8 |
| Region-based Methods | Faster [11] | | | | | 16.8 |
| | DA-Faster [52] | √ | | | | 32.2 |
| | G-L-Faster [51] | √ | √ | | | 35.9 |

*2) Limitations of Pixel-level Adaptation Methods:* Existing pixel-level adaptive detectors [42]-[45] are not conducive to the rapid deployment of the model from the source domain to the target domain, since it requires long time to train a CycleGAN based model to generate images that look similar to those from the target domain. More importantly, it can be observed from Fig. 5 that, for the cross-posture hidden object detection where the layout of the image content changes significantly between domains, the CycleGAN based model cannot reduce such a cross-domain discrepancy in the pixel space.

*3) Cross-posture Hidden Object Detection Using DSEMs:* When adapting the detector from the hands-down posture to the hands-up posture, due to the poor quality of target images generated by the pixel-level adaptation, we have to employ the feature-level adaptation to perform cross-posture hidden object detection for MMW images. We conduct the experiments and report the results of both some common feature-level adaptation methods [51], [52] and our DSEMs in Table VI. It can be seen that our method greatly improves the capability of

TABLE VII
ABLATION STUDY OF DOMAIN ADAPTIVE FOREGROUND ENHANCEMENT AND DENSELY-SEMANTIC ALIGNMENT IN DSEM FOR ADAPTING CONV4_3 OF VGG-16. FE AND DA DENOTE THE FOREGROUND ENHANCEMENT AND THE DOMAIN ADAPTATION UNIT, RESPECTIVELY. W/O MEANS WITHOUT.

| Method | ORI | $l=0$ with FE | $l=0$ w/o DA | $l=0$ w/o FE | $l=1$ with FE | $l=1$ w/o DA | $l=1$ w/o FE | $l=3$ with FE | $l=3$ w/o DA | $l=3$ w/o FE | $l=4$ with FE | $l=4$ w/o DA | $l=4$ w/o FE |
|---|---|---|---|---|---|---|---|---|---|---|---|---|---|
| SSD+DSEM | 27.6 | **33.6** | 31.5 | 31.0 | **35.8** | 33.9 | 33.3 | **38.1** | 37.1 | 36.6 | **37.1** | 35.4 | 36.1 |
| RDet+DSEM | 22.8 | **27.7** | 26.4 | 25.7 | **29.3** | 28.4 | 28.1 | **34.7** | 33.9 | 33.8 | **34.0** | 33.8 | 33.5 |

TABLE VIII
FOREGROUND/BACKGROUND SEGMENTATION RESULTS FOR THE DOMAIN ADAPTIVE FOREGROUND ENHANCEMENT ON CLIPART DATASET.

| Segmentation Network | mIoU |
|---|---|
| FE w/o DA | 58.4 |
| FE | 69.2 |

TABLE IX
ABLATION STUDY OF MULTIPLE POOLING BRANCHES IN DSEMS. P-1248 DENOTES THE POOLED FEATURES WITH SPATIAL SIZES OF 1×1, 2×2, 4×4, 8×8.

| Method | mAP |
|---|---|
| SSD [14] | 27.6 |
| SSD+DSEMs w/o P | 38.0 |
| SSD+DSEMs with P-124 | 39.1 |
| SSD+DSEMs with P-1248 | **40.1** |
| RDet [16] | 22.8 |
| RDet+DSEMs w/o P | 34.8 |
| RDet+DSEMs with P-124 | 35.2 |
| RDet+DSEMs with P-1248 | **36.7** |

TABLE X
ABLATION STUDY OF ALIGNING DIFFERENT LAYERS USING DSEMS. THE DEFINITION OF DN-8, DN-16, *ETC*, FOLLOWS TABLE I.

| Method | DN-8 | DN-16 | DN-32 | DN-64 | DN-100 | DN-300 | mAP |
|---|---|---|---|---|---|---|---|
| SSD+DSEMs | ✓ | ✓ | | | | | 39.5 |
| | ✓ | | ✓ | | | | **40.1** |
| | ✓ | | | ✓ | | | 37.7 |
| | ✓ | | | | ✓ | | 35.9 |
| | ✓ | | | | | ✓ | 36.6 |
| RDet+DSEMs | ✓ | ✓ | | | \ | \ | **36.7** |
| | ✓ | | ✓ | | \ | \ | 36.3 |
| | ✓ | | | ✓ | \ | \ | 35.9 |

TABLE XI
RESULTS ON ADAPTATION FROM PASCAL VOC TO CLIPART FOR APPLYING DSEM TO FASTER RCNN.

| Method | Backbone | mAP |
|---|---|---|
| Faster RCNN [11] | VGG16 | 23.5 |
| G-L-Faster [51] | VGG16 | 31.5 |
| Faster RCNN+DSEM (ours) | VGG16 | **33.1** |

TABLE XII
RESULTS ON ADAPTATION FROM PASCAL VOC TO CLIPART **ONLY 3** UNLABELED IMAGES PER CLASS FROM THE TARGET DOMAIN.

| Method | Backbone | mAP |
|---|---|---|
| SSD [14] | VGG16 | 27.6 |
| G-L-SSD [51] | VGG16 | 28.5 |
| SSD+DSEMs (ours) | VGG16 | **33.2** |

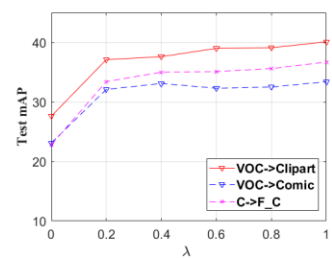

Fig. 6. Sensitivity analyses for $\lambda$. VOC, C and F_C denote PASCAL VOC, Cityscapes and FoggyCityscapes, respectively.

cross-domain detection and outperforms the DA-Faster [52] and G-L-Faster [51] with a considerable margin.

## IV. INSIGHT ANALYSES

### A. Ablation Studies

We conduct the ablation studies from four aspects to evaluate the impact of each designed module in DSEM on the mAP value. They include: 1) the domain adaptive foreground enhancement, 2) the densely-connected representations, 3) the multi-scale feature encoder, and 4) adapting different layers of region-free detectors.

For the first item, as shown in Table VII, the mAP achieved by the proposed method with FE outperforms the proposed method without FE. This is mainly because FE can suppress more background information and enable the network to focus on the alignment of foreground regions. Besides, the foreground enhancement with the domain adaptation unit outperforms the foreground enhancement without the domain adaptation unit, indicating that the domain adaptation unit in the $F_{seg}$ plays an important role in cross-domain detection of both SSD and RefineDet. Further, in Table VIII, we show the mIoU of foreground/background segmentation for the foreground enhancement.

For the second item, Table VII also shows that with the increase of $l$, the domain adaptability of the SSD and RefineDet gradually improves. We consider that the improvement of adaptability mainly results from the increasing capability of the DSEM to encode multiple-level semantic features from the backbone network.

For the third item, Table IX indicates that the mAP is changing when employing pooling kernels with different sizes.

For the fourth item, we report the results of aligning different layers of SSD and RefineDet in Table X. The results show that aligning the high-level features degrades the performance since these low-resolution features lose more spatial information.

### B. Faster RCNN with DSEM

To comprehensively investigate the effectiveness of DSEM on the region-based detectors, we conduct the experiments of applying the DSEM to Faster RCNN. Specifically, we first train

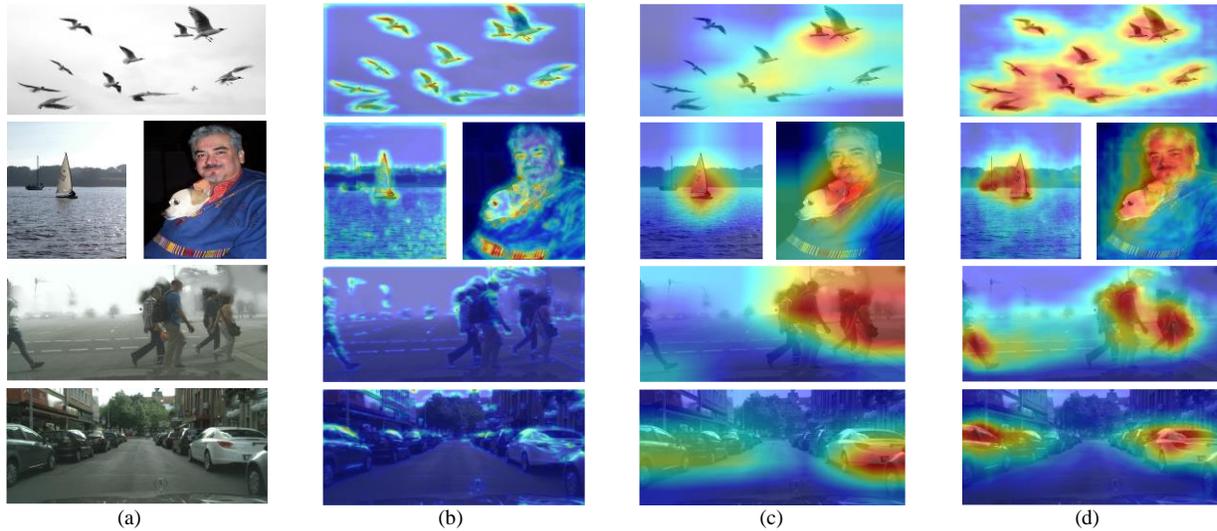

Fig. 7. Visualization of domain evidence using Grad-CAM [58]. (a) Input images. (b) Domain evidence obtained by L-SSD (the local alignment module in [49], [51]). (c) Domain evidence obtained by G-SSD (the global alignment module in [49], [51]). (d) Domain evidence obtained by our DSEM.

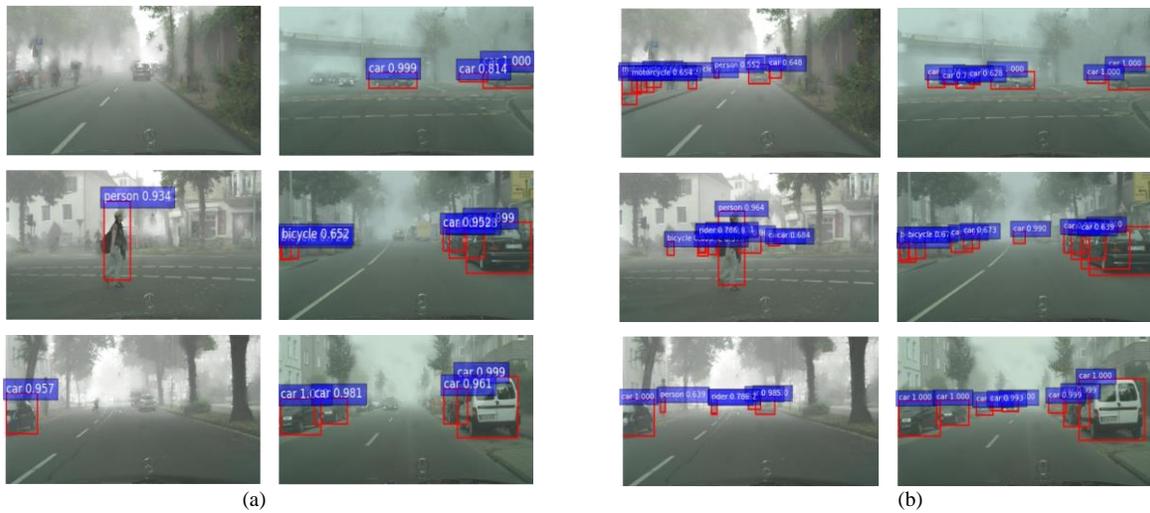

Fig. 8. Visual detection examples for adapting RefineDet and RefineDet+DSEMs from Cityscapes to FoggyCityscapes. (a) Detection results predicted by RefineDet. (b) Detection results predicted by RefineDet+DSEMs.

Faster RCNN on the labeled source images, according to the official training setups [11]. Next, we insert the DSEM into Faster RCNN to align the Conv4_3 layer, and jointly fine-tune the pre-trained source model on both the labeled source images (PASCAL VOC) and the unlabeled target images (Clipart) to alleviate the inter-domain discrepancy. The fine-tuning process performs 3k iterations in total.

Table XI shows that the proposed DSEM is pluggable into Faster RCNN and beneficial to boost its cross-domain detection performance, verifying the effectiveness and generality of our method.

### C. Sensitivity Analyses and Model Interpretability

We first conduct the sensitivity study of $\lambda$ in Eq. (9). Fig. 6 shows that our method achieves the best results when $\lambda$ equals to 1. Then, we give a good visual explanation by means of Grad-CAM [58], illustrating which features or image regions are fed into the domain classifier to perform the alignment process. The Grad-CAM is a novel localization technique, which can make DNNs based model more transparent and interpretable. We use the Grad-CAM to visualize the domain-related evidence (heatmaps) represented by our DSEM in Fig. 7. The visualization results show that compared with typical domain adaptation modules, DSEM can capture more important regions and the encoded features are semantically rich. Thus, the backbone detection network can focus on multiple instances to deceive the domain classifier. By comparing Fig. 7(b) and (d), it can be observed that although domain-related features learned by the local alignment module [49], [51] focus on the instances, they mainly include low-level information, such as features that describe edges and textures. In contrast, our DSEM can learn more semantically-rich features of instances.

### D. Visual Detection Examples and Failure Cases

We show some visual detection results for adapting the region-free detector from Cityscapes to FoggyCityscapes in Fig. 8. Further, Fig. 9 illustrates failure cases for cross-domain models. The most common failure cases come from the objects with the similar appearances in the semantic features, such as guideboards and pedestrians from the target domain.

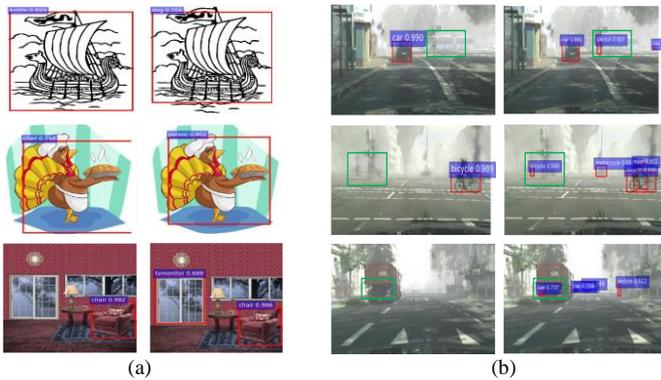

Fig. 9. Failure cases on (a) Clipart and (b) FoggyCityscapes for the baseline models (left) and the corresponding proposed models (right), where green bounding boxes highlight mispredicted objects.

*E. Few-shot Cross-domain Object Detection*

For UDA, a large number of unlabeled images from the target domain are still needed. It should be more practical to adapt the detector only using a few target images. Thus, we conduct the experiments that only a few images (here, only 3 unlabeled target images per class) from the target domain are given. For few-shot cross-domain object detection, we use the same training setups as UDA. It can be seen from Table XII that the cross-domain detection accuracy of the proposed method (*i.e.,* SSD+DSEMs) decreases from 40.1% to 33.2%, when only 3 unlabeled target images per class are given. This implies that the domain adaptive detector adapted to the biased target distribution formed by only a few target samples, which needs further study in the future.

## V. CONCLUSION

In this paper, we have developed a domain adaptive one-stage region-free framework and further proposed an adversarial module, densely semantic enhancement module (DSEM), which is pluggable into various region-free detectors to boost the domain adaptability of the region-free detector family. Firstly, to enhance the cross-domain matching of some important image regions, the DSEM learns to predict a transferable foreground mask, which can be utilized to suppress the background regions. Secondly, considering that the region-free detector family recognizes objects of different sizes using features from different layers, the DSEM encodes multi-scale representations between different domains. Extensive experiment results and insightful analyses have demonstrated that the DSEM not only captures some crucial image regions to achieve the feature-level alignment, improving the domain adaptability of region-free detectors, but also outperforms the existing domain adaptive region-based detector such as G-L-Faster RCNN.

Since training images from the target domain are usually difficult to collect in some real cross-domain scenarios, future work would focus on studying the few-shot cross-domain detection problem, where only a few annotated images from the target domain are given. Additionally, considering that the performance on the target domain may be a key clue to adapt the source domain detector to the target domain, cross-domain object detection using reward strategy will be a possible research topic in our future work.